\documentclass{article}

\usepackage{microtype}
\usepackage{graphicx}
\usepackage{booktabs} 

\usepackage{hyperref}

\usepackage{algorithm}
\usepackage[noend]{algpseudocode}
\usepackage{subcaption}
\usepackage[textsize=tiny]{todonotes}

\usepackage{amsmath,amsfonts,bm}

\def\eqref#1{equation~\ref{#1}}

\def\1{\bm{1}}

\DeclareMathAlphabet{\mathsfit}{\encodingdefault}{\sfdefault}{m}{sl}
\SetMathAlphabet{\mathsfit}{bold}{\encodingdefault}{\sfdefault}{bx}{n}

\makeatletter
\newcounter{algorithmbis}
\def\BState{\State\hskip-\ALG@thistlm}
\renewcommand{\thealgorithmbis}{\arabic{algorithmbis}}
\def\algorithmbis{\@ifnextchar[{\@algorithmbisa}{\@algorithmbisb}}
\def\@algorithmbisa[#1]{%
  \refstepcounter{algorithmbis}
  \trivlist
  \leftmargin\z@
  \itemindent\z@
  \labelsep\z@
  \item[\parbox{\textwidth}{%
    \hrule
    \hrule
    \noindent\strut\textbf{Algorithm \thealgorithmbis} #1
    \hrule
  }]\hfil\vskip0em%
}
\def\@algorithmbisb{\@algorithmbisa[]}

\makeatother

\usepackage[accepted]{icml2021}

\icmltitlerunning{Detecting and Quantifying Malicious Activity with Simulation-based Inference}

\begin{document}

\twocolumn[
\icmltitle{Detecting and Quantifying Malicious Activity with Simulation-based Inference}



\icmlsetsymbol{equal}{*}

\begin{icmlauthorlist}
\icmlauthor{Andrew Gambardella}{ox}
\icmlauthor{Bogdan State}{sc}
\icmlauthor{Naeemullah Khan}{ox}
\icmlauthor{Leo Tsourides}{fb}
\icmlauthor{Philip H. S. Torr}{ox}
\icmlauthor{Atılım Güneş Baydin}{ox}
\end{icmlauthorlist}

\icmlaffiliation{ox}{Department of Engineering Science, University of Oxford}
\icmlaffiliation{fb}{Facebook Inc.}
\icmlaffiliation{sc}{scie.nz, work done while at Facebook}

\icmlcorrespondingauthor{Andrew Gambardella}{gambs@robots.ox.ac.uk}

\icmlkeywords{Machine Learning, ICML}

\vskip 0.3in
]



\printAffiliationsAndNotice{}  

\begin{abstract}
We propose the use of probabilistic programming techniques to tackle the malicious user identification problem in a recommendation algorithm. Probabilistic programming provides numerous advantages over other techniques, including but not limited to providing a disentangled representation of how malicious users acted under a structured model, as well as allowing for the quantification of damage caused by malicious users. We show experiments in malicious user identification using a model of regular and malicious users interacting with a simple recommendation algorithm, and provide a novel simulation-based measure for quantifying the effects of a user or group of users on its dynamics.
\end{abstract}

\section{Introduction}

In 1993 a famous New Yorker cartoon of a computer-browsing canine dryly proclaimed, ``on the Internet nobody knows you're a dog.'' With hindsight this ur-meme has proven prescient 
with respect to the problem of authenticity on the Internet. That any one Internet user can have identities that are both multitudinous and mutable formed an important part of the network's promise as a medium for communication, self-expression and empowerment. But flexible identities also carry with them the risk of deception, with Internet-facilitated fraud coming to cast a dark shadow over the luminous future originally envisaged by techno-optimists \citep{friedman2001social}.

Stakes increase dramatically when the problem of authenticity meets the power of ranking algorithms, which are responsible for fulfilling and defining information retrieval needs. Tricking an algorithm into honoring at face value those features coming from a certain set of (malicious) users can result in disaster, with unsuspecting users being recommended content, the consumption of which serves purely to enrich a set of attackers rather than to fulfill users' information needs.

Ideally speaking, a good recommendations system should be able to identify and remove malicious users before they can disrupt the ranking system by a significant margin. However, to eliminate the risk of false positives a resilient ranking system can use as much data as possible. So we have to adjust the tradeoff between false positives and the damage a set of malicious users can cause to a ranking system. 


Bearing these limitations in mind, as a first approximation, it seems reasonable, for the sake of greater analytical clarity, to divide the user base of an online social network into the vast majority of organic users and a minority of attack profiles. Seen in this way, discussions of inauthentic amplification and coordinated inauthentic behavior fit in with earlier analyses of ``shilling'' in recommender systems \citep{si2020shilling}. Here, we study a popular class of shilling attacks known as profile injection attacks \cite{Williams2007}, in which attackers add bogus accounts to a recommendation system, and attempt to push the ratings of a certain subset of products upward and others downward, while obfuscating their intentions \cite{Ricci2015}.

In our case we are interested in asking the deceptively simple question, \textit{how would ranking outcomes differ in the absence of malicious users}. We define malicious users here as those users misrepresenting either their motives or their identity for strategic gain involving the promotion or demotion of units of content. The question is difficult to answer because of the multitudinous feedback loops potentially at work in recommender systems, which mean that simply counting the effects directly attributable to known malicious users is insufficient for giving an accurate picture of ranking outcomes in the putative counterfactual universe in which no malicious users existed.

Probabilistic programming \citep{van2018introduction} has emerged as a principled means of dealing with complex causal scenarios not unlike the issue discussed here, being used in domains as diverse as lion behavior interpretation \citep{Dhir2017}, spacecraft trajectories \citep{acciarini-2020-spacecraft} and high-energy physics \citep{Baydin2018,Baydin2019}. It is our contention that probabilistic programming, and simulation-based inference \citep{cranmer2020frontier} in general, can be used credibly to estimate the difference between realized outcomes and the counter-factual scenario which excludes malicious behavior. We support our assertion by providing:
\begin{enumerate}
    \item A proof-of-concept detection algorithm, validating that malicious user identification using simulation-based inference techniques is possible using data from the model. This is a necessary but insufficient condition for the computation of a counterfactual scenario.
    \item A simulation-based counterfactual measure of influence in a ranking algorithm, grounded in an information theoretical view of the joint probability distributions of ranking outcomes in the presence, and absence, of malicious users.
    \item An illustration of how the measure could be applied -- we show that malicious users acting in coordination have greater impact on social network dynamics than those acting independently.

\end{enumerate}


\section{Related Work}

The misuse of recommendation algorithms for adversarial gain dates back to the Internet's first growth spurt as a communication platform, and is inherently intertwined with the history of digital spam, defined as:
\begin{quote}
``the attempt to abuse of, or manipulate, a techno-social system by producing and injecting unsolicited, and/or undesired content aimed at steering the behavior of humans or the system itself, at the direct or indirect, immediate or long-term advantage of the spammer(s).'' \citep{ferrara2019history}
\end{quote}

Recommendation algorithms have been a key vector for the amplification of spam since the 1990s, when automated -- rather than curated -- information retrieval first became feasible in a consumer setting, thanks to  \citeauthor{Page1998}'s (1999) now-famous development of the PageRank algorithm. Compared to earlier proposals, PageRank notably provided a mechanism which enforced algorithmic resiliency -- a recursive definition of popularity which protected against simplistic attempts at faking site popularity for monetary gain through the construction of hyperlinking rings (``spamdexing'').

PageRank became the algorithmic foundation for Google, the dominant search engine of the past two decades. Nonetheless, in what would become a common pattern in the Internet industry, its original mechanisms proved only partially effective against adaptive adversaries. The rise of ranking also gave birth to an entire industry, search engine optimization (SEO), dedicated to improving results against the ranking algorithm, sometimes using adversarial ``black hat'' methods \citep{malaga2010search}. This evolution, in turn, led to subsequent changes to Google's algorithms to improve their resiliency against adversaries \citep{mccullagh2011testing}. 



Adversarial attacks against recommendation algorithms have become increasingly prominent recently, given the importance of social media in shaping contentious news cycles rife with misinformation, in particular during the course of events such as the 2016 U.S. general elections \citep{allcott2017social}, or the 2018 Brazilian general elections \citep{machado2019study}. Automated posting and engagement, via ``social bots'' has been recognized as a particularly important factor in the spread of disinformation on social media \citep{ferrara2016rise, arnaudo2017computational, shao2018spread, cresci2019detecting}. The issue of automation is intertwined with that of inauthenticity, with ``coordinated inauthentic behavior'' (CIB) emerging as a distinct concept both among academics \citep{giglietto2020takes} and industry practitioners \citep{weedon2017information}.

The concept of CIB relies on the existence of ``inauthentic accounts,'' distinguishable from authentic users. The notion of authenticity carries with it a great deal of complexity on the Internet. Our analysis as a result is scoped to those platforms (such as Facebook or Twitter) which rely on the explicit expectation of online identities consistent with offline personas. Even CIB itself should not be seen simply through the binary view of malicious attackers and organic users, as attackers may act to catalyze existing grievances and ideologies present among audiences ripe for manipulation \citep{starbird2019disinformation}.

To fight vote spam in user--item interactions, \citet{bian2008few} proposed training ranking models using a method based on simulated voting spam at training time. Alternatively, \citet{bhattacharjee2007algorithms} have proposed creating incentives for power users (\textit{``connoisseurs''}) to counter-act the influence of spammers. More recently, \citet{basat2017game} have proposed introducing noise into the ranking function to account for distorted incentives leading to the production of low-quality content (e.g., through link farming).


\section{Methods}

Our examination is meant to provide a minimal example of a recommender system. We choose movie ranking as our setting, given the canonical nature of the task, e.g., IMDb\footnote{\url{https://www.imdb.com/}} data having a long history of use in the study of recommender systems. This is admittedly a ``toy'' model, which does not account for more complex designs (i.e., personalization), or for 
the many issues that intervene in the deployment of recommender systems in the real world (model update cycles, A/B testing, site outages, etc.). Formulating and studying such a model allows us to focus on the derivation of core concepts such as the influence metric (Section~\ref{sec:influence}) in the framework of probabilistic programming.

We created a model that represents several malicious users attempting to game a recommendation algorithm modeled as a simple ranking algorithm (without loss of generality, users rating movies and being recommended new movies to watch based on the current mean rating), provided in the supplementary material. In this model, multiple users (some malicious and some organic) are rating items, which are then ranked and suggested to other users based on their ranking. User tastes are modeled by real-valued variables $\nu_i \in [0, 1]$, which determine which movies they would naturally like. Similarly, movies have taste features $\mu_j \in [0, 1]$ which denote something akin to their genre and in our model are left fixed. We model the rating function $\mathrm{Rate}(\upsilon_i, \mu_j)$ so that user $i$ will rate movie $j$ higher the closer user taste $\nu_i$ is to movie taste $\mu_j$. The resulting ratings $\rho_{i, j}$ in each user--movie pair constitute the elements of the global rating matrix $\bm{R}$.


In this model the main latent variables we would like to infer are the binary variables $\beta_i$, denoting whether a given user $i$ is malicious or not, and $\tau_i$, denoting the target movie which user $i$ would like to boost, if user $i$ is malicious, i.e., if $\beta_i = 1$. Probabilistic programming will allow us to condition this model on a given rating matrix (for instance, one that represents real-world movie ratings), and then find empirical distributions over the latent variables in the simulator ($\bm{\mu}, \bm{\nu}, \bm{\beta}, \bm{\tau}$) consistent with the given rating matrix $\bm{R_\textrm{obs}}$. In summary, for the purposes of identifying malicious users and what they are trying to boost, we will obtain the posterior distribution $p(\bm{\beta}, \bm{\tau}\vert\bm{R_\textrm{obs}})$, while leaving $\bm{\mu}$ and $\bm{\nu}$ as nuisance variables.


We implemented our model in PyProb \citep{Baydin2019}, a lightweight probabilistic programming library for stochastic simulators. We obtain our posteriors using weighted importance sampling \cite{Kitagawa1996} which gives a posterior in the form of weighted traces drawn from a proposal distribution, which is in our case the unmodified stochastic simulator. As our posterior is given to us in the form of simulations conditioned on observed data, it is by nature completely disentangled and interpretable, and will tell us the goals of each of the malicious users ($\tau_i$) in addition to their identities ($\beta_i$). Crucially, this Bayesian approach also gives us principled uncertainty estimates associated with all our predictions.


\section{Experiments}

\subsection{Obtaining the Posterior and Identifying Malicious Users}

We found that obtaining a posterior over the identities of malicious users, as well as their targets, was non-trivial, with different inference engine families behaving considerably differently. Markov-chain Monte Carlo (MCMC) \cite{Metropolis1953, Hastings1970, wingate2011lightweight}, despite being the gold standard to converge to the correct posterior given enough samples, performed extremely poorly. We suspect that this is due to the variables of interest (the malicious users and their targets) making up only a small portion of the latent variables in the model, as well as being discrete whereas the others are continuous. We observed that the model as it is currently formulated was not a good fit for the single-site MCMC \citep{wingate2011lightweight, van2018introduction} inference engine implemented in PyProb, mainly because generating proposals where a single maliciousness latent $\beta_i$ is flipped lead to very low acceptance probabilities due to the very abrupt nature of the resulting change in the rating matrix (which needs to be compensated by corresponding changes in some user taste latents $\upsilon_i$ that cannot be achieved in a single-site MCMC algorithm), leading to slow mixing and poor sample efficiency. 

We found that weighted importance sampling (IS) \cite{Kitagawa1996} performed better in practice than MCMC, and used it to obtain posteriors conditioned on observed ratings matrices. When using IS with 100,000 executions of our model in which malicious users do not attempt to disguise their activities (i.e., difficulty $\alpha = 0$), the mean of rating matrices in the posterior, i.e., the posterior predictive $p(\bm{R}|\bm{R}_{\mathrm{obs}})$, appears to be a noisy version of the observed ground truth rating matrix $\bm{R}_{\mathrm{obs}}$, showing that the inference scheme sampled a posterior distribution over simulation runs in which the observed rating matrix is likely.

Much more interesting are scenarios in which malicious users attempt to disguise their activities through obfuscated attacks. We model this obfuscation by setting the difficulty hyperparameter $\alpha=0.3$. This obfuscation introduces a significant amount of uncertainty into our results, which is reflected in the detection of malicious users. Whereas in the unambiguous case IS produced an empirical posterior over malicious users and malicious target that matched with the ground truth nearly exactly, in the ambiguous case we see significant uncertainty in the empirical posterior. The results show that the most probable (0.85) explanation of the observed rating matrix is that there are no malicious users, although we also see non-negligible (0.15) probability attached to the scenario where there is one malicious user (which we know to be the ground truth for the observed rating matrix). We also see that the mode of the posterior distribution for the malicious users matches the ground truth (user 4), albeit with much lower probability (0.15) compared with the unambiguous case (1.0). We also see that while there is probability mass associated with the ground truth value of the malicious target, this probability is quite low, showing that the IS inference scheme has significant uncertainty in the identification of the malicious target, given the experimental setup and the number of traces (100,000) that we ran during inference.

\subsection{Using Counterfactuals to Quantify the Damage Done by Malicious Users}
\label{sec:influence}

\begin{figure}
    \centering
    \includegraphics[scale=0.55]{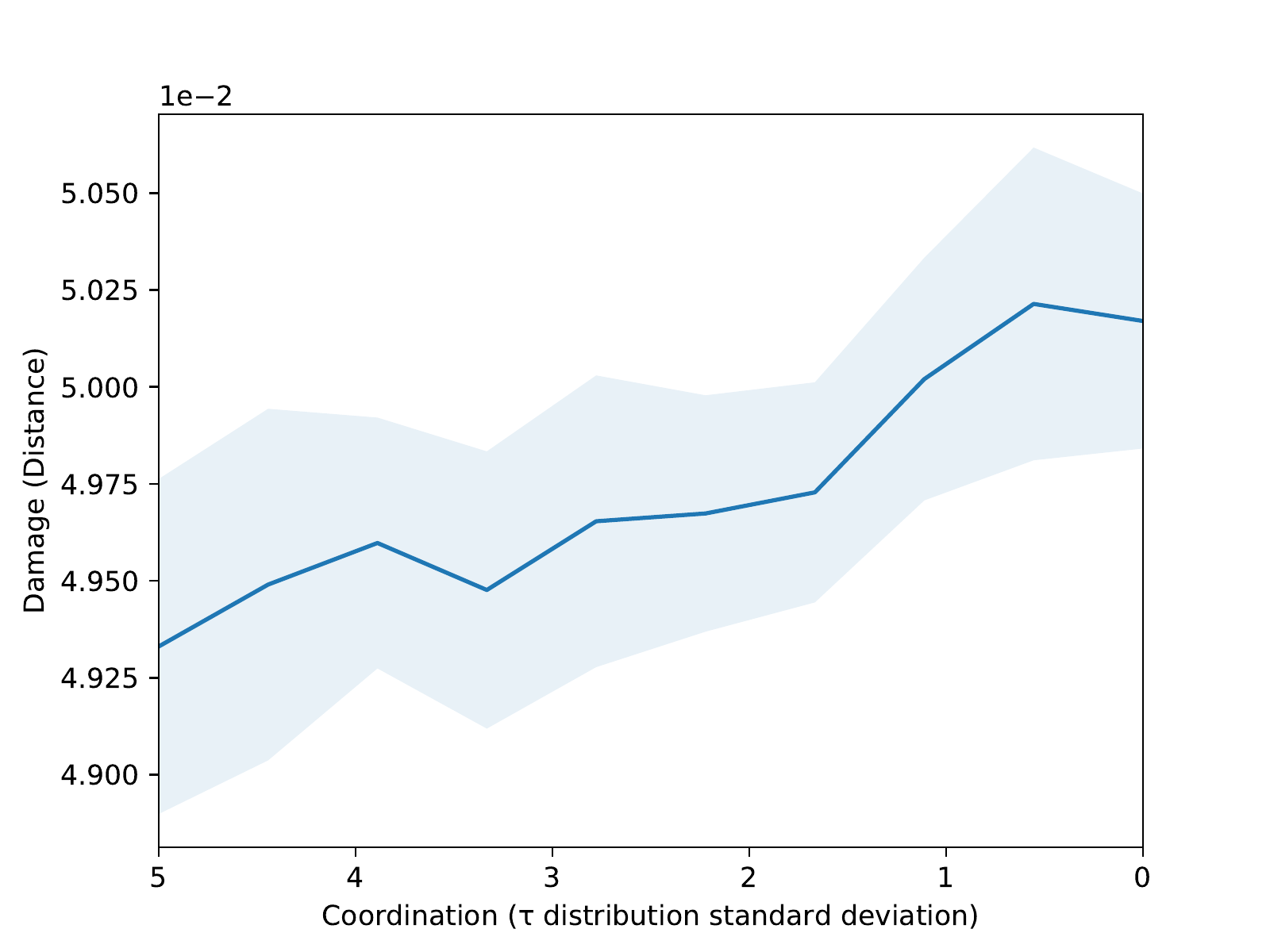}
    \caption{Measuring the effect on the rating matrix as the standard deviation of the distribution from which we draw $\tau$ is increased. Lower values of standard deviation (i.e., more coordination between malicious actors) results in a higher impact to the dynamics of the ratings in the matrix. Results averaged over 10 different seeds, with one standard deviation bounds shown.}
    \label{fig:influence}
\end{figure}

In addition to giving disentangled and interpretable explanations of the behaviour of users interacting with a ranking algorithm, simulation-based inference also gives us the ability to measure the effects of users on the model's dynamics. We can quantify the amount of impact that users imposed on the dynamics of an observed rating matrix by using a slightly modified model, which allows us to disarm users by nullifying the effect of their ratings. Comparing the dynamics reflected in the distributions both with and without disarmed users gives us a counterfactual-based method of quantifying their impact on the dynamics of the simulator. Given a set of disarmed users $\bm{\gamma}$, we find the distance between the posterior-predictive distributions $p(\bm{R}\vert\bm{R}_\mathrm{obs})$ and $p(\bm{R}\vert\bm{R}_\mathrm{obs},\bm{\gamma})$ given observed data $\bm{R}_\mathrm{obs}$, or between the prior-predictive $p(\bm{R})$ and $p(\bm{R}\vert\bm{\gamma})$ in the generic case without an observed $\bm{R}_{\mathrm{obs}}$.

Our chosen metric for measuring the impact that a disarmed user or group of users has caused is the average JS distance \citep{Endres2003}, the square root of the symmetric JS divergence \citep{Dagan1997}, computed as the average of JS distances between the counterfactual and real probability distributions over ratings, per entry in the rating matrix. In a distribution over rating matrices $p(\bm{R}|\cdot)$, each entry in the matrix is a probability distribution (in the empirical case, a histogram) over ratings for each user--movie pair over a large number of simulator executions. Our measure for impact is then the average JS distance for each of these histograms, between the realized and counterfactual rating matrices. The JS distance is a valid metric between probability distributions and always normalised to $[0,1]$, making it an attractive choice to measure distances between distributions over matrix entries.

We show results using our influence measure between counterfactual and realised ratings matrices while varying the amount of coordination between malicious users in Figure \ref{fig:influence}. When malicious targets are drawn from a distribution with a low standard deviation, malicious actors act in a more coordinated fashion (as they are more likely to target the same movie), which leads to a higher average distance between the counterfactual and realized ratings distributions. As the malicious target standard deviation increases, malicious users act against each others' interests, leading to a lower overall impact on the dynamics of the model.

\section{Discussion}

We have suggested the use of probabilistic programming techniques to both discover and measure the influence of malicious users interacting with a ranking algorithm. We base our choice of this method on its conceptual advantages in modeling the full universe of possibilities deriving from the complex interactions between users, content and ranking algorithms. The use of simulation-based approaches in this setting has been limited by practical concerns, until recent advances in numerical computation -- coupled with the emergence of high-quality libraries for probabilistic programming.

Probabilistic programming techniques also have some important additional advantages over other methods. They provide for interpretable explanations as to, e.g., why a user would be classified as malicious, and provide measures of confidence in their predictions. Furthermore, generative modeling of the entities and processes involved in this setting allows us to make precise definitions of the core concepts and to quantify key aspects such as user influence and malicious user damage.




\bibliography{nash}

\begin{thebibliography}{32}
\providecommand{\natexlab}[1]{#1}
\providecommand{\url}[1]{\texttt{#1}}
\expandafter\ifx\csname urlstyle\endcsname\relax
  \providecommand{\doi}[1]{doi: #1}\else
  \providecommand{\doi}{doi: \begingroup \urlstyle{rm}\Url}\fi

\bibitem[Acciarini et~al.(2020)Acciarini, Pinto, Metz, Boufelja, Kaczmarek,
  Merz, Martinez-Heras, Letizia, Bridges, and
  Baydin]{acciarini-2020-spacecraft}
Acciarini, G., Pinto, F., Metz, S., Boufelja, S., Kaczmarek, S., Merz, K.,
  Martinez-Heras, J.~A., Letizia, F., Bridges, C., and Baydin, A.~G.
\newblock {Spacecraft Collision Risk Assessment with Probabilistic
  Programming}.
\newblock In \emph{Third Workshop on Machine Learning and the Physical Sciences
  (NeurIPS 2020), Vancouver, Canada}, 2020.

\bibitem[Allcott \& Gentzkow(2017)Allcott and Gentzkow]{allcott2017social}
Allcott, H. and Gentzkow, M.
\newblock {Social media and fake news in the 2016 election}.
\newblock \emph{Journal of Economic Perspectives}, 31\penalty0 (2):\penalty0
  211--236, 2017.
\newblock ISSN 08953309.
\newblock \doi{10.1257/jep.31.2.211}.

\bibitem[Arnaudo(2017)]{arnaudo2017computational}
Arnaudo, D.
\newblock {Computational Propaganda in Brazil: Social Bots during Elections}.
\newblock \emph{Computational Propaganda Research Project}, 8:\penalty0 1--39,
  2017.

\bibitem[Basat et~al.(2017)Basat, Tennenholtz, and Kurland]{basat2017game}
Basat, R.~B., Tennenholtz, M., and Kurland, O.
\newblock {A game theoretic analysis of the adversarial retrieval setting}.
\newblock \emph{Journal of Artificial Intelligence Research}, 60:\penalty0
  1127--1164, 2017.
\newblock ISSN 10769757.
\newblock \doi{10.1613/jair.5547}.

\bibitem[Baydin et~al.(2019{\natexlab{a}})Baydin, Heinrich, Bhimji,
  Gram-Hansen, Louppe, Shao, Prabhat, Cranmer, and Wood]{Baydin2018}
Baydin, A.~G., Heinrich, L., Bhimji, W., Gram-Hansen, B., Louppe, G., Shao, L.,
  Prabhat, P., Cranmer, K., and Wood, F.
\newblock {Efficient probabilistic inference in the quest for physics beyond
  the standard model}.
\newblock In \emph{Advances in Neural Information Processing Systems},
  volume~33, 2019{\natexlab{a}}.

\bibitem[Baydin et~al.(2019{\natexlab{b}})Baydin, Shao, Bhimji, Heinrich,
  Meadows, Liu, Munk, Naderiparizi, Gram-Hansen, Louppe, Ma, Zhao, Torr, Lee,
  Cranmer, Prabhat, and Wood]{Baydin2019}
Baydin, A.~G., Shao, L., Bhimji, W., Heinrich, L., Meadows, L., Liu, J., Munk,
  A., Naderiparizi, S., Gram-Hansen, B., Louppe, G., Ma, M., Zhao, X., Torr,
  P., Lee, V., Cranmer, K., Prabhat, and Wood, F.
\newblock {Etalumis: Bringing probabilistic programming to scientific
  simulators at scale}.
\newblock In \emph{International Conference for High Performance Computing,
  Networking, Storage and Analysis, SC}, 2019{\natexlab{b}}.
\newblock ISBN 9781450362290.
\newblock \doi{10.1145/3295500.3356180}.

\bibitem[Bhattacharjee \& Goel(2007)Bhattacharjee and
  Goel]{bhattacharjee2007algorithms}
Bhattacharjee, R. and Goel, A.
\newblock {Algorithms and incentives for robust ranking}.
\newblock In \emph{Proceedings of the Annual ACM-SIAM Symposium on Discrete
  Algorithms}, volume 07-09-Janu, pp.\  425--433, 2007.
\newblock ISBN 9780898716245.

\bibitem[Bian et~al.(2008)Bian, Agichtein, Liu, and Zha]{bian2008few}
Bian, J., Agichtein, E., Liu, Y., and Zha, H.
\newblock {A few bad votes too many? Towards robust ranking in social media}.
\newblock In \emph{AIRWeb 2008 - Proceedings of the 4th International Workshop
  on Adversarial Information Retrieval on the Web}, pp.\  53--60, 2008.
\newblock ISBN 9781605581590.
\newblock \doi{10.1145/1451983.1451997}.

\bibitem[Cranmer et~al.(2020)Cranmer, Brehmer, and Louppe]{cranmer2020frontier}
Cranmer, K., Brehmer, J., and Louppe, G.
\newblock {The frontier of simulation-based inference}.
\newblock \emph{Proceedings of the National Academy of Sciences}, 117\penalty0
  (48):\penalty0 30055--30062, 2020.
\newblock ISSN 0027-8424.
\newblock \doi{10.1073/pnas.1912789117}.

\bibitem[Cresci(2020)]{cresci2019detecting}
Cresci, S.
\newblock {Detecting malicious social bots: Story of a never-ending clash}.
\newblock In \emph{Lecture Notes in Computer Science (including subseries
  Lecture Notes in Artificial Intelligence and Lecture Notes in
  Bioinformatics)}, volume 12021 LNCS, pp.\  77--88. Springer, 2020.
\newblock ISBN 9783030396268.
\newblock \doi{10.1007/978-3-030-39627-5_7}.

\bibitem[Dagan et~al.(1997)Dagan, Lee, and Pereira]{Dagan1997}
Dagan, I., Lee, L., and Pereira, F.
\newblock {Similarity-based methods for word sense disambiguation}.
\newblock In \emph{Proceedings of the Thirty-Fifth Annual Meeting of the
  Association for Computational Linguistics and Eighth Conference of the
  European Chapter of the Association for Computational Linguistics}, pp.\
  56--63, 1997.
\newblock \doi{10.3115/979617.979625}.

\bibitem[Dhir et~al.(2017)Dhir, Wood, V{\'{a}}k{\'{a}}r, Markham, Wijers,
  Trethowan, {Du Preez}, Loveridge, and MacDonald]{Dhir2017}
Dhir, N., Wood, F., V{\'{a}}k{\'{a}}r, M., Markham, A., Wijers, M., Trethowan,
  P., {Du Preez}, B., Loveridge, A., and MacDonald, D.
\newblock {Interpreting lion behaviour with nonparametric probabilistic
  programs}.
\newblock In \emph{Proceedings of the Conference on Uncertainty in Artificial
  Intelligence (UAI)}, 2017.

\bibitem[Endres \& Schindelin(2003)Endres and Schindelin]{Endres2003}
Endres, D.~M. and Schindelin, J.~E.
\newblock {A new metric for probability distributions}.
\newblock \emph{IEEE Transactions on Information Theory}, 49\penalty0
  (7):\penalty0 1858--1860, 2003.
\newblock ISSN 00189448.
\newblock \doi{10.1109/TIT.2003.813506}.

\bibitem[Ferrara(2019)]{ferrara2019history}
Ferrara, E.
\newblock {The History of Digital Spam}.
\newblock \emph{Communications of the ACM}, 62\penalty0 (8):\penalty0 82--91,
  2019.

\bibitem[Ferrara et~al.(2016)Ferrara, Varol, Davis, Menczer, and
  Flammini]{ferrara2016rise}
Ferrara, E., Varol, O., Davis, C., Menczer, F., and Flammini, A.
\newblock {The rise of social bots}.
\newblock \emph{Communications of the ACM}, 59\penalty0 (7):\penalty0 96--104,
  2016.
\newblock ISSN 15577317.
\newblock \doi{10.1145/2818717}.

\bibitem[Friedman \& Resnick(2001)Friedman and Resnick]{friedman2001social}
Friedman, E.~J. and Resnick, P.
\newblock {The social cost of cheap pseudonyms}.
\newblock \emph{Journal of Economics and Management Strategy}, 10\penalty0
  (2):\penalty0 173--199, 2001.
\newblock ISSN 10586407.
\newblock \doi{10.1162/105864001300122476}.

\bibitem[Giglietto et~al.(2020)Giglietto, Righetti, Rossi, and
  Marino]{giglietto2020takes}
Giglietto, F., Righetti, N., Rossi, L., and Marino, G.
\newblock {It takes a village to manipulate the media: coordinated link sharing
  behavior during 2018 and 2019 Italian elections}.
\newblock \emph{Information Communication and Society}, 23\penalty0
  (6):\penalty0 867--891, 2020.
\newblock ISSN 14684462.
\newblock \doi{10.1080/1369118X.2020.1739732}.

\bibitem[Hastings(1970)]{Hastings1970}
Hastings, W.~K.
\newblock {Monte Carlo Sampling Methods Using Markov Chains and Their
  Applications}.
\newblock \emph{Biometrika}, 57\penalty0 (1):\penalty0 97, 1970.
\newblock ISSN 00063444.
\newblock \doi{10.2307/2334940}.

\bibitem[Kitagawa(1996)]{Kitagawa1996}
Kitagawa, G.
\newblock {Monte Carlo Filter and Smoother for Non-Gaussian Nonlinear State
  Space Models}.
\newblock \emph{Journal of Computational and Graphical Statistics}, 1996.
\newblock ISSN 10618600.
\newblock \doi{10.2307/1390750}.

\bibitem[Machado et~al.(2019)Machado, Kira, Narayanan, Kollanyi, and
  Howard]{machado2019study}
Machado, C., Kira, B., Narayanan, V., Kollanyi, B., and Howard, P.~N.
\newblock {A study of misinformation in whatsapp groups with a focus on the
  brazilian presidential elections}.
\newblock In \emph{The Web Conference 2019 - Companion of the World Wide Web
  Conference, WWW 2019}, pp.\  1013--1019, 2019.
\newblock ISBN 9781450366755.
\newblock \doi{10.1145/3308560.3316738}.

\bibitem[Malaga(2010)]{malaga2010search}
Malaga, R.~A.
\newblock {Search Engine Optimization—Black and White Hat Approaches}.
\newblock In \emph{Advances in Computers}, volume~78, pp.\  1--39. Elsevier,
  2010.
\newblock \doi{10.1016/s0065-2458(10)78001-3}.

\bibitem[McCullagh(2011)]{mccullagh2011testing}
McCullagh, D.
\newblock {Testing Google's Panda Algorithm: CNET Analysis}.
\newblock
  \url{https://www.cnet.com/news/testing-googles-panda-algorithm-cnet-analysis/},
  2011.
\newblock URL \url{http://news.cnet.com/8301-31921_3-20054797-281.html}.

\bibitem[Metropolis et~al.(1953)Metropolis, Rosenbluth, Rosenbluth, Teller, and
  Teller]{Metropolis1953}
Metropolis, N., Rosenbluth, A.~W., Rosenbluth, M.~N., Teller, A.~H., and
  Teller, E.
\newblock {Equation of state calculations by fast computing machines}.
\newblock \emph{Journal of Chemical Physics}, 21\penalty0 (6):\penalty0
  1087--1092, 1953.
\newblock ISSN 00219606.
\newblock \doi{10.1063/1.1699114}.

\bibitem[Page et~al.(1998)Page, Brin, Motwani, and Winograd]{Page1998}
Page, L., Brin, S., Motwani, R., and Winograd, T.
\newblock {The PageRank Citation Ranking: Bringing Order to the Web}.
\newblock \emph{World Wide Web Internet And Web Information Systems}, 1998.
\newblock ISSN 1752-0509.
\newblock \doi{10.1.1.31.1768}.

\bibitem[Ricci et~al.(2015)Ricci, Shapira, and Rokach]{Ricci2015}
Ricci, F., Shapira, B., and Rokach, L.
\newblock \emph{{Recommender systems handbook, Second edition}}.
\newblock 2015.
\newblock ISBN 9781489976376.
\newblock \doi{10.1007/978-1-4899-7637-6}.

\bibitem[Shao et~al.(2017)Shao, Ciampaglia, Varol, Yang, Flammini, and
  Menczer]{shao2018spread}
Shao, C., Ciampaglia, G.~L., Varol, O., Yang, K., Flammini, A., and Menczer, F.
\newblock {The spread of low-credibility content by social bots}.
\newblock \emph{Nature communications}, 9\penalty0 (1):\penalty0 1--9, 2017.

\bibitem[Si \& Li(2020)Si and Li]{si2020shilling}
Si, M. and Li, Q.
\newblock {Shilling attacks against collaborative recommender systems: a
  review}.
\newblock \emph{Artificial Intelligence Review}, 53\penalty0 (1):\penalty0
  291--319, 2020.
\newblock ISSN 15737462.
\newblock \doi{10.1007/s10462-018-9655-x}.

\bibitem[Starbird et~al.(2019)Starbird, Arif, and
  Wilson]{starbird2019disinformation}
Starbird, K., Arif, A., and Wilson, T.
\newblock {Disinformation as collaborative work: Surfacing the participatory
  nature of strategic information operations}.
\newblock \emph{Proceedings of the ACM on Human-Computer Interaction},
  3\penalty0 (CSCW):\penalty0 1--26, 2019.
\newblock ISSN 25730142.
\newblock \doi{10.1145/3359229}.

\bibitem[van~de Meent et~al.(2018)van~de Meent, Paige, Yang, and
  Wood]{van2018introduction}
van~de Meent, J.-W., Paige, B., Yang, H., and Wood, F.
\newblock {An Introduction to Probabilistic Programming}.
\newblock \emph{arXiv preprint}, 2018.

\bibitem[Weedon et~al.(2017)Weedon, Nuland, and Stamos]{weedon2017information}
Weedon, J., Nuland, W., and Stamos, A.
\newblock {Information Operations and Facebook}, 2017.
\newblock ISSN 1047-9651.

\bibitem[Williams et~al.(2007)Williams, Mobasher, and Burke]{Williams2007}
Williams, C.~A., Mobasher, B., and Burke, R.
\newblock {Defending recommender systems: Detection of profile injection
  attacks}.
\newblock \emph{Service Oriented Computing and Applications}, 2007.
\newblock ISSN 18632386.
\newblock \doi{10.1007/s11761-007-0013-0}.

\bibitem[Wingate et~al.(2011)Wingate, Stuhlm{\"{u}}ller, and
  Goodman]{wingate2011lightweight}
Wingate, D., Stuhlm{\"{u}}ller, A., and Goodman, N.~D.
\newblock {Lightweight implementations of probabilistic programming languages
  via transformational compilation}.
\newblock In \emph{Journal of Machine Learning Research}, volume~15, pp.\
  770--778, 2011.

\end{thebibliography}
\bibliographystyle{icml2021}

\appendix
\onecolumn
\section{Recommender Model}
\begin{algorithm}[H]
    \caption{Probabilistic generative model of movie ratings based on the movie ranking setting, defining the joint distribution $p(\bm{\mu}, \bm{\upsilon}, \bm{\beta}, \bm{\tau}, \bm{R}) = p(\bm{R}\vert\bm{\mu}, \bm{\upsilon}, \bm{\beta}, \bm{\tau})\,p(\bm{\mu}, \bm{\upsilon}, \bm{\beta}, \bm{\tau})$, where $p(\bm{R}\vert \cdot)$ is the likelihood and $p(\bm{\mu}, \bm{\upsilon}, \bm{\beta}, \bm{\tau})$ is the prior. Given an observed rating matrix $\bm{R}_\textrm{obs}$, we would like to infer the posterior $p(\bm{\beta}, \bm{\tau}\vert\bm{R_\textrm{obs}}).$}
    \label{alg:imdb1}

    \begin{flushleft}
    \hspace*{\algorithmicindent} $N_\mu\in\mathbb{N}$: Number of movies\\
    \hspace*{\algorithmicindent} $N_\upsilon\in\mathbb{N}$: Number of users\\
    \hspace*{\algorithmicindent} $p_\beta\in[0,1]$: Probability of maliciousness\\
    \hspace*{\algorithmicindent} $\rho_\beta\in[0,1]$: Mean malicious rating\\
    \hspace*{\algorithmicindent} $\rho_\sigma\in\mathbb{R}^{+}$: Rating standard deviation\\
    \hspace*{\algorithmicindent} $\tau_\mu\in[0,N_\mu]$: Mean malicious target\\
    \hspace*{\algorithmicindent} $\tau_\sigma\in\mathbb{R}^{+}$: Malicious target standard deviation\\
    \hspace*{\algorithmicindent} $T\in\mathbb{N}$: Time steps\\
    \hspace*{\algorithmicindent} $\alpha\in[0,1]$: Difficulty\\
    \end{flushleft}
    \begin{algorithmic}[1]
    \Procedure{RatingModel}{}
	\BState \emph{Sample latents}:
    
    \State $\bm{\mu} \gets$ fixed vector, components sampled only once as $\{\mu_i \sim \mathrm{Uniform}(0, 1)\}_{i=1}^{N_\mu}$\Comment Movies and taste features
    \State $\bm{\upsilon} \gets \{\upsilon_i \sim \mathrm{Uniform}(0, 1)\}_{i=1}^{N_\upsilon}$\Comment Users and taste features
    \State $\bm{\beta} \gets \{\beta_i \sim \mathrm{Bernoulli}(p_\beta)\}_{i=1}^{N_\upsilon}$\Comment Maliciousness of users
    \State $\bm{\tau} \gets \{\tau_i \sim \mathrm{TruncatedNormal}(\tau_\mu, \tau_\sigma, 0, N_\mu)\}_{i=1}^{N_\upsilon}$\Comment Maliciousness targets (ignored for organic users)

	\BState \emph{Simulate}:
    \State $\bm{R} \gets \{\rho_{i,j} = 0\}_{i=1, j=1}^{N_\upsilon, N_\mu}$\Comment Rating matrix ($N_\upsilon$ rows, $N_\mu$ cols)
    
    \For{$t=1,\dots,T$}
    	\For{$i=1,\dots,N_\upsilon$}
    		\If{$\beta_i$}\Comment{Malicious user}
    		    \If{$\tau_i$ is unrated}
    			    \State $j \gets \tau_i$ \Comment Pick malicious target
    			    \State $\rho \gets (1-\alpha)*\rho_\beta + \alpha*\mathrm{Rate}(\upsilon_i, \mu_j)$ \Comment Rate to boost $\tau_i$
    			\Else
    			    \State $j \gets \mathrm{Pick}(\bm{R})$ \Comment Pick movie based on ranking
    			    \State $\rho \gets \alpha*\mathrm{Rate}(\upsilon_i, \mu_j)$ \Comment Rate lower than or equal to taste feature match
    			\EndIf
    		\Else\Comment{Organic user}
    			\State $j \gets \mathrm{Pick}(\bm{R})$ \Comment Pick movie based on ranking
    			\State $\rho \gets \mathrm{Rate}(\upsilon_i, \mu_j)$ \Comment Rate according to taste feature match
    		\EndIf
    		\State $\rho_{i, j} \sim \mathrm{TruncatedNormal}(\rho, \rho_\sigma, 0, 1)$\Comment Sample rating by user $i$ for movie $j$
    	\EndFor
    \EndFor
    \State {\textbf return} $\bm{R}$\Comment Return rating matrix

    \EndProcedure

    \Procedure{Pick}{$\bm{R}$}
        \State {$\bm{m} \gets \{m_j = \frac{1}{N_{\upsilon}}\sum_{i=1}^{N_{\upsilon}}{\rho_{i, j}} \}_{j=1}^{N_{\mu}}$}\Comment Mean movie ratings
        \State $j \sim \mathrm{Categorical}(N_{\mu}, \bm{m})$\Comment Sample movie index based on mean ratings
        \State {\textbf return} $j$\Comment Return picked movie index j/
    \EndProcedure

    \Procedure{Rate}{$\upsilon_i, \mu_j$}
        \State {\textbf return} $1 - \left| \upsilon_i - \mu_j \right|$\Comment Return rating
    \EndProcedure

    \end{algorithmic}
\end{algorithm}

\newpage
\section{Conditioning Results, Unambiguous Problem}
\begin{figure}[H]
    \centering
    \begin{subfigure}{0.49\textwidth}
    \includegraphics[scale=0.3]{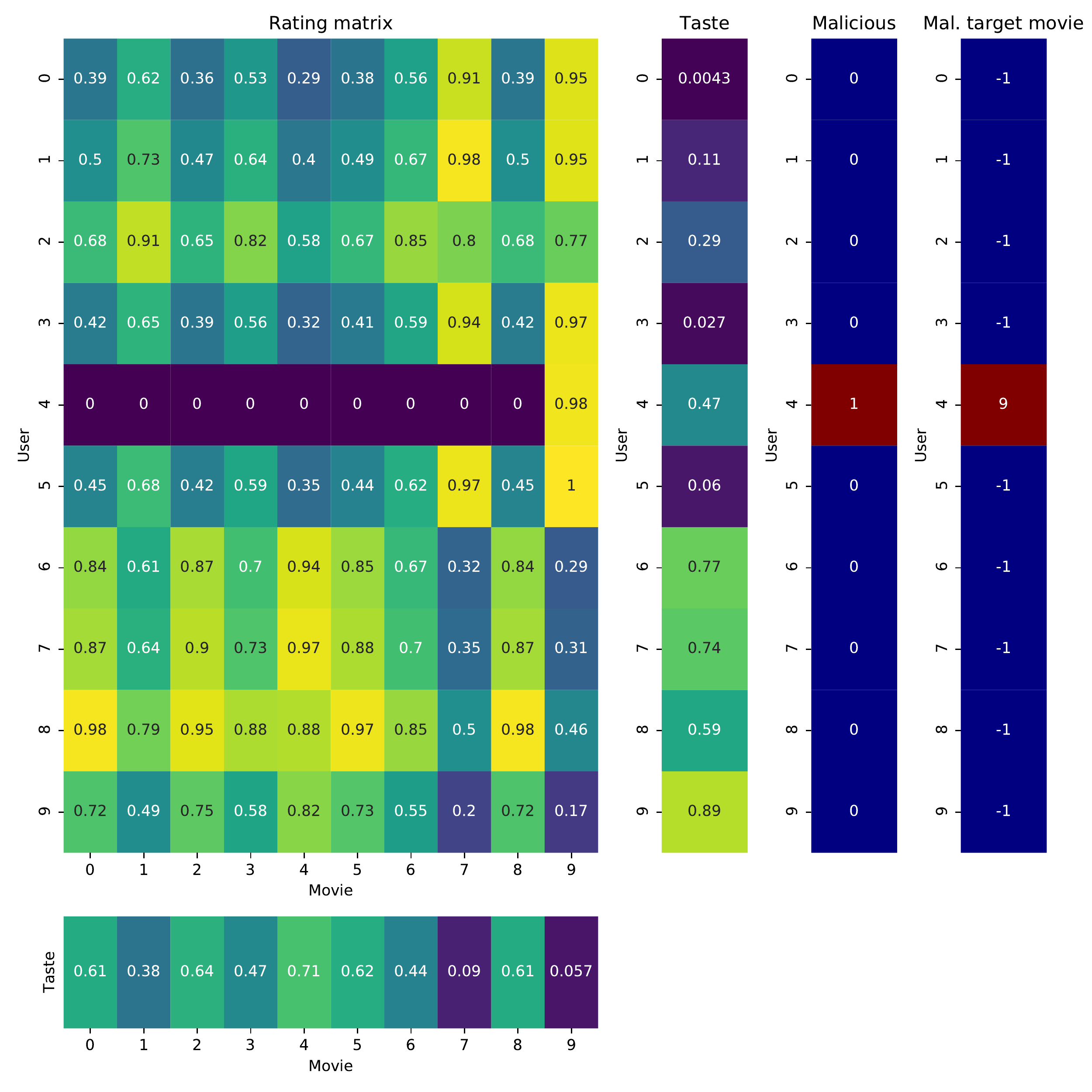}
    \caption{Observed rating matrix $\bm{R}_{\mathrm{obs}}$}
    \end{subfigure}
    \begin{subfigure}{0.49\textwidth}
    \includegraphics[scale=0.3]{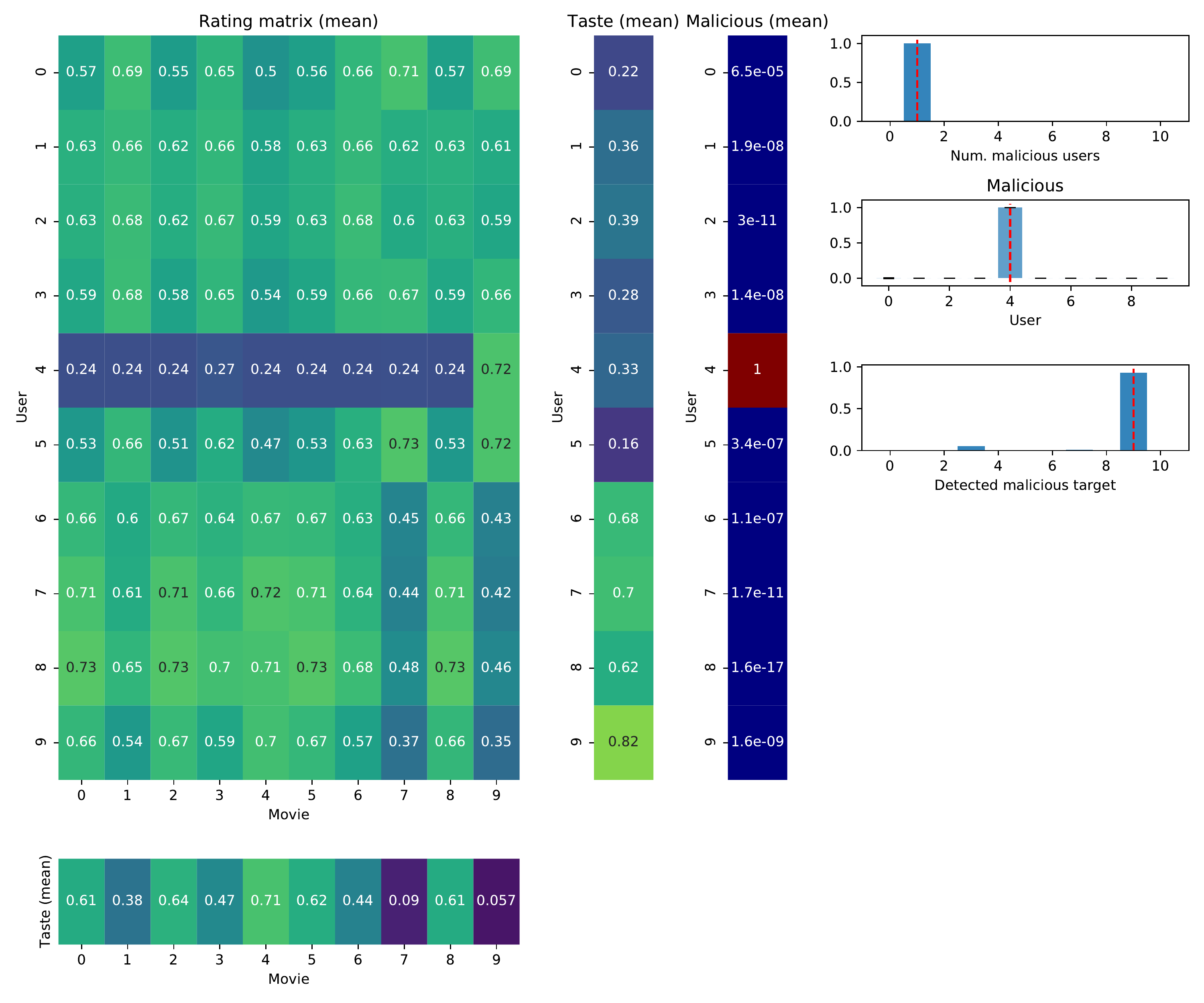}
    \caption{Empirical posterior $p(\bm{\beta},\bm{\tau},\bm{R}|\bm{R}_{\mathrm{obs}})$}
    \end{subfigure}
    \caption{(a) Rating matrix $\bm{R}_{\mathrm{obs}}$ that we observe as the input, and the corresponding ground truth values for user maliciousness $\bm{\beta}$ and targets $\bm{\tau}$. (b) Empirical posterior distribution found by weighted IS in a scenario in which malicious users do not attempt to disguise their activities ($\alpha=0$). We can see that the posterior predictive rating matrix, $p(\bm{R}|\bm{R}_{\mathrm{obs}})$ matches fairly closely with the ground truth $\bm{R}_{\mathrm{obs}}$, and that the posterior malicious users $p(\bm{\beta}, \bm{\tau}|\bm{R}_{\mathrm{obs}})$ match up with the ground truth in all posterior simulations. IS is able to discover both the malicious user and its target with little uncertainty. Dashed vertical lines show ground truth values.}
    \label{fig:gt_post_comparison_unambiguous}
\end{figure}

\newpage
\section{Conditioning Results, Ambiguous Problem}
\begin{figure}[H]
    \centering
    \begin{subfigure}{0.49\textwidth}
    \includegraphics[scale=0.3]{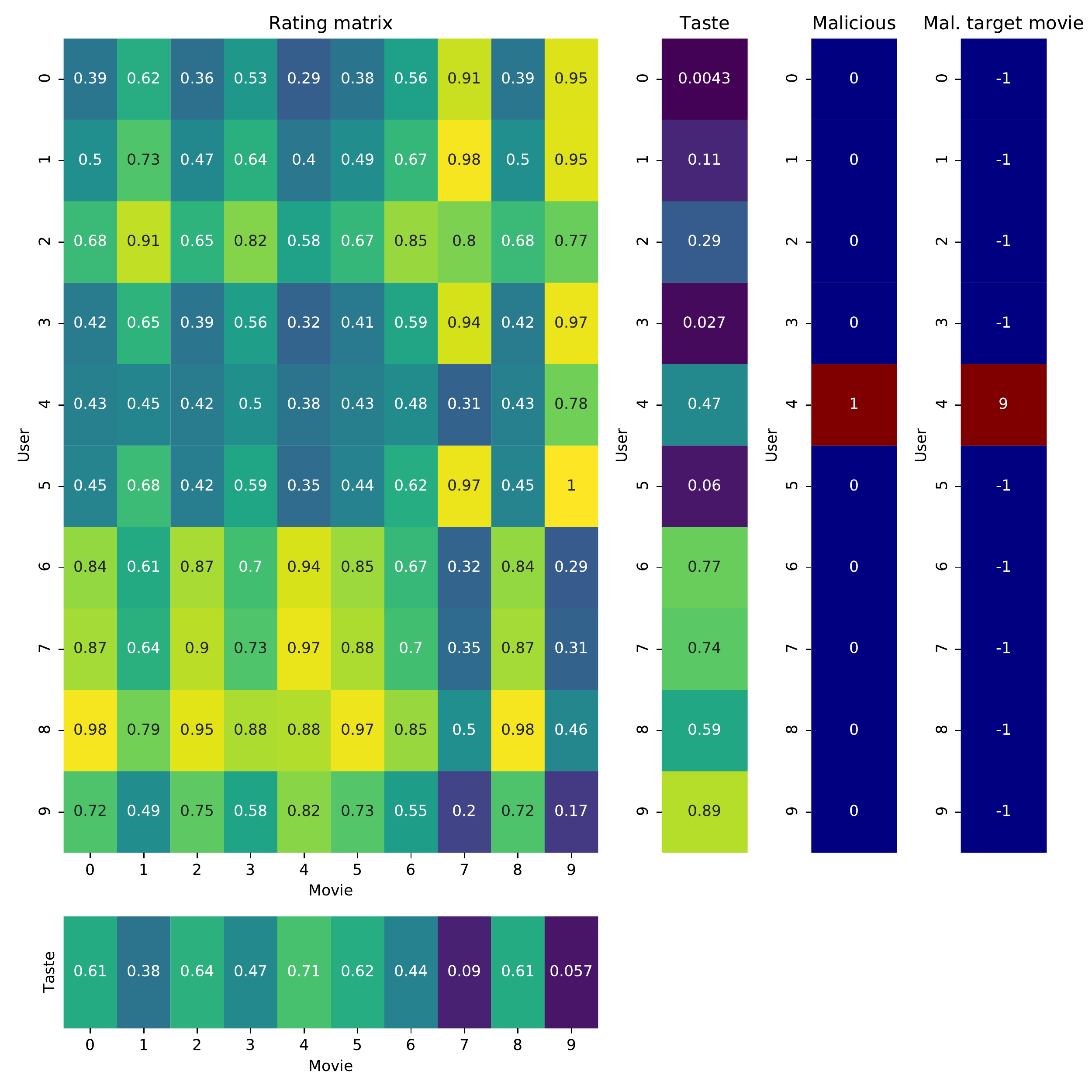}
    \caption{Observed rating matrix $\bm{R}_{\mathrm{obs}}$}
    \end{subfigure}
    \begin{subfigure}{0.49\textwidth}
    \includegraphics[scale=0.3]{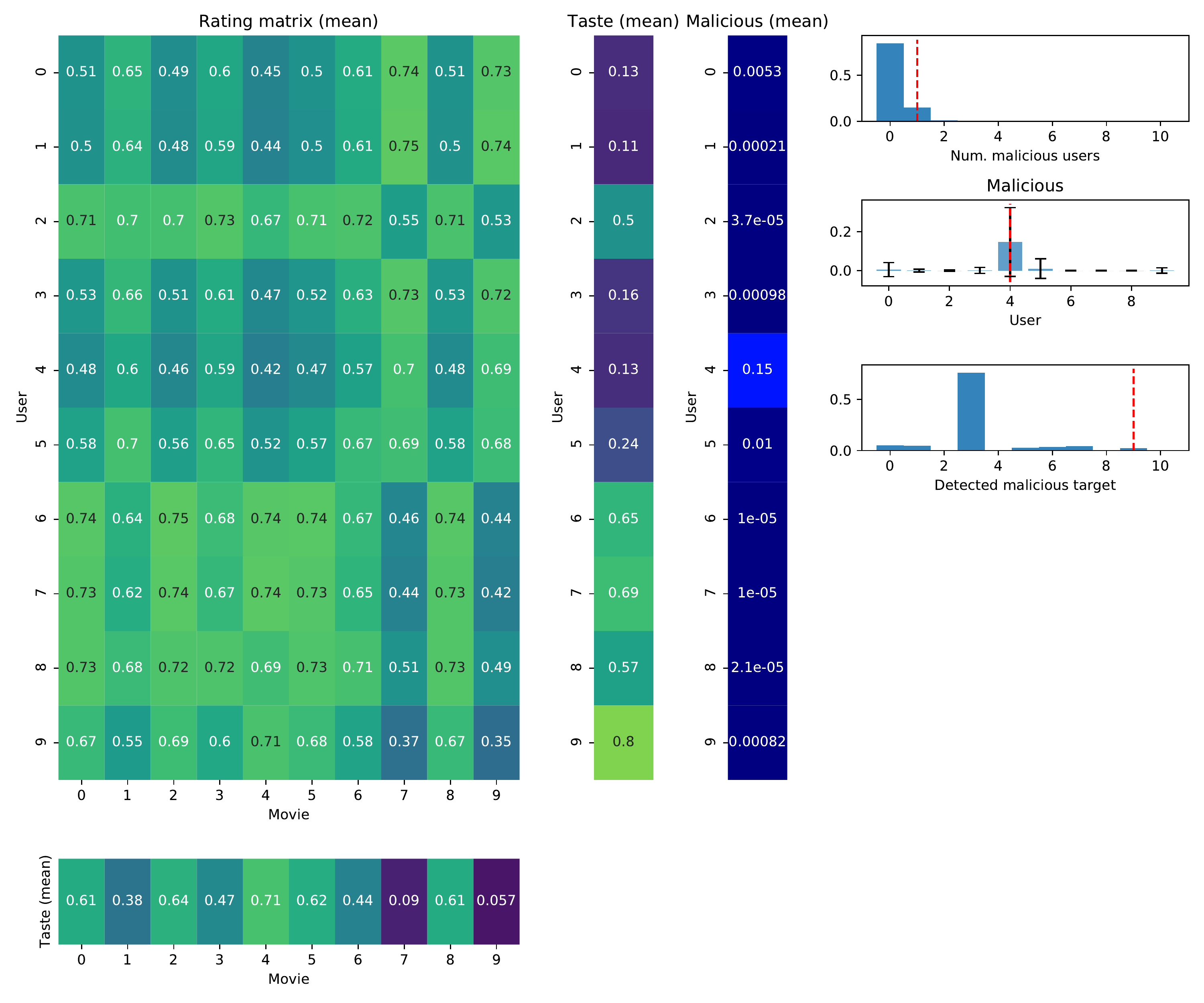}
    \caption{Empirical posterior $p(\bm{\beta},\bm{\tau},\bm{R}|\bm{R}_{\mathrm{obs}})$}
    \end{subfigure}
    \caption{(a) Rating matrix $\bm{R}_{\mathrm{obs}}$ that we observe as the input, sampled from a scenario in which malicious imitate non-malicious users ($\alpha=0.3$). (b) Empirical posterior found by weighted IS. IS is able to discover this malicious user in about 15\% of simulations making up the empirical posterior, but cannot discover their target accurately. Dashed vertical lines show ground truth values.}
    \label{fig:gt_post_comparison_ambiguous}
\end{figure}

\newpage
\section{Algorithm with Disarmed Users}
\nopagebreak
\begin{algorithm}[H]
    \caption{Variation of Algorithm~\ref{alg:imdb1}, in which we can disarm a subset of users $\bm{\gamma}$ and quantify the effect (influence) of one or more users in the rating distribution, e.g., by computing a distance between predictive distributions $p(\bm{R})$ and $p(\bm{R}\vert\bm{\gamma})$ for the prior, or $p(\bm{R}\vert\bm{R}_\mathrm{obs})$ and $p(\bm{R}\vert\bm{R}_\mathrm{obs},\bm{\gamma})$ for the posterior conditioned on a given rating matrix observation $\bm{R}_{\mathrm{obs}}$.}
    \label{alg:imdb2}

    \begin{flushleft}
    \hspace*{\algorithmicindent} $N_\mu\in\mathbb{N}$: Number of movies\\
    \hspace*{\algorithmicindent} $N_\upsilon\in\mathbb{N}$: Number of users\\
    \hspace*{\algorithmicindent} $p_\beta\in[0,1]$: Probability of maliciousness\\
    \hspace*{\algorithmicindent} $\rho_\beta\in[0,1]$: Mean malicious rating\\
    \hspace*{\algorithmicindent} $\rho_\sigma\in\mathbb{R}^{+}$: Rating standard deviation\\
    \hspace*{\algorithmicindent} $\tau_\mu\in[0,N_\mu]$: Mean malicious target\\
    \hspace*{\algorithmicindent} $\tau_\sigma\in\mathbb{R}^{+}$: Malicious target standard deviation\\
    \hspace*{\algorithmicindent} $T\in\mathbb{N}$: Time steps\\
    \hspace*{\algorithmicindent} $\alpha\in[0,1]$: Difficulty\\
    \hspace*{\algorithmicindent} $\bm{\gamma}: \{\gamma_i \in {0, 1}\}_{i=1}^{N_\upsilon}$: Disarmed users
    \end{flushleft}
    \begin{algorithmic}[1]
    \Procedure{RatingModel}{}
	\BState \emph{Sample latents}:
    
    \State $\bm{\mu} \gets$ fixed vector, components sampled only once as $\{\mu_i \sim \mathrm{Uniform}(0, 1)\}_{i=1}^{N_\mu}$\Comment Movies and taste features
    \State $\bm{\upsilon} \gets \{\upsilon_i \sim \mathrm{Uniform}(0, 1)\}_{i=1}^{N_\upsilon}$\Comment Users and taste features
    \State $\bm{\beta} \gets \{\beta_i \sim \mathrm{Bernoulli}(p_\beta)\}_{i=1}^{N_\upsilon}$\Comment Maliciousness of users
    \State $\bm{\tau} \gets \{\tau_i \sim \mathrm{TruncatedNormal}(\tau_\mu, \tau_\sigma, 0, N_\mu)\}_{i=1}^{N_\upsilon}$\Comment Maliciousness targets (ignored for organic users)

	\BState \emph{Simulate}:
    \State $\bm{R} \gets \{\rho_{i,j} = 0\}_{i=1, j=1}^{N_\upsilon, N_\mu}$\Comment Rating matrix ($N_\upsilon$ rows, $N_\mu$ cols)
    
    \For{$t=1,\dots,T$}
    	\For{$i=1,\dots,N_\upsilon$}
    	\If{not $\gamma_i$} \Comment Do not let user rate if disarmed
    		\If{$\beta_i$}\Comment{Malicious user}
    		    \If{$\tau_i$ is unrated}
    			    \State $j \gets \tau_i$ \Comment Pick malicious target
    			    \State $\rho \gets (1-\alpha)*\rho_\beta + \alpha*\mathrm{Rate}(\upsilon_i, \mu_j)$ \Comment Rate to boost $\tau_i$
    			\Else
    			    \State $j \gets \mathrm{Pick}(\bm{R})$ \Comment Pick movie based on ranking
    			    \State $\rho \gets \alpha*\mathrm{Rate}(\upsilon_i, \mu_j)$ \Comment Rate lower than or equal to taste feature match
    			\EndIf
    		\Else\Comment{Organic user}
    			\State $j \gets \mathrm{Pick}(\bm{R})$ \Comment Pick movie based on ranking
    			\State $\rho \gets \mathrm{Rate}(\upsilon_i, \mu_j)$ \Comment Rate according to taste feature match
    		\EndIf
    		\State $\rho_{i, j} \sim \mathrm{TruncatedNormal}(\rho, \rho_\sigma, 0, 1)$\Comment Sample rating by user $i$ for movie $j$
        \EndIf
    	\EndFor
    \EndFor
    \State {\textbf return} $\bm{R}$\Comment Return rating matrix

    \EndProcedure

    \Procedure{Pick}{$\bm{R}$}
        \State {$\bm{m} \gets \{m_j = \frac{1}{N_{\upsilon}}\sum_{i=1}^{N_{\upsilon}}{\rho_{i, j}} \}_{j=1}^{N_{\mu}}$}\Comment Mean movie ratings
        \State $j \sim \mathrm{Categorical}(N_{\mu}, \bm{m})$\Comment Sample movie index based on mean ratings
        \State {\textbf return} $j$\Comment Return picked movie index j/
    \EndProcedure

    \Procedure{Rate}{$\upsilon_i, \mu_j$}
        \State {\textbf return} $1 - \left| \upsilon_i - \mu_j \right|$\Comment Return rating
    \EndProcedure

    \end{algorithmic}
\end{algorithm}



\end{document}